\documentclass{article}
\usepackage{spconf,amsmath,graphicx}
\usepackage{algorithmic,algorithm}
\usepackage{amssymb}
\usepackage{amsmath}
\usepackage{bm}
\newtheorem{definition}{Definition}

\title{RESTRICTED CONNECTION ORTHOGONAL MATCHING PURSUIT \\FOR SPARSE SUBSPACE CLUSTERING}
\name{Wenqi Zhu\qquad Yuesheng Zhu*\qquad Li Zhong\qquad Shuai Yang\thanks{*Corresponding author. This work is supported by the Shenzhen Municipal Development and Reform Commission (Disciplinary Development Program for Data Science and Intelligent Computing), and by Shenzhen International cooperative research projects GJHZ20170313150021171,and in part by NSFC-Shenzhen Robot Jointed Founding (U1613215).Contact us: wenqizhu@pku.edu.cn, zhuys@pku.edu.cn.}}
\address{Communication and Information Security Lab, Institute of Big Data Technologies,\\Shenzhen Graduate School, Peking University}
\begin{document}
\setlength{\textfloatsep}{7pt}
\setlength{\abovedisplayskip}{0pt}
\setlength{\belowdisplayskip}{0pt}
\setlength{\floatsep}{5pt}
\setlength{\abovecaptionskip}{0pt}
\setlength{\belowcaptionskip}{0pt}
\topmargin = 0mm

\maketitle
\begin{abstract}

Sparse Subspace Clustering (SSC) is one of the most popular methods for clustering data points into their underlying subspaces. However, SSC may suffer from heavy computational burden. Orthogonal Matching Pursuit applied on SSC accelerates the computation but the trade-off is the loss of clustering accuracy. In this paper, we propose a noise-robust algorithm, Restricted Connection Orthogonal Matching Pursuit for Sparse Subspace Clustering (RCOMP-SSC), to improve the clustering accuracy and maintain the low computational time by restricting the number of connections of each data point during the iteration of OMP. Also, we develop a framework of control matrix to realize RCOMP-SCC. And the framework is scalable for other data point selection strategies. Our analysis and experiments on synthetic data and two real-world databases (EYaleB \& Usps) demonstrate the superiority of our algorithm compared with other clustering methods in terms of accuracy and computational time.
\end{abstract}
\begin{keywords}
Restricted Connection, Noise Robust, Orthogonal Matching Pursuit (OMP), Sparse Subspace Clustering (SSC)
\end{keywords}
\section{Introduction}
\label{sec:intro}

The unsupervised learning methods such as Subspace Clustering(SC) \cite{2} that can learn from the unlabeled data have become more and more important. SC can be applied in many computer vision tasks in which high dimensional data can be approximated as a union of low dimensional subspaces \cite{2,3}, including face clustering \cite{4}, image representation \cite{5}, motion segmentation \cite{6}, and written digit clustering \cite{7}.

There are several kinds of SC methods, including algebraic, iterative, statistical, and spectral clustering-based methods \cite{2}. Further the spectral clustering-based methods are divided into different categories according to the norm regularization they choose for sparse coefficients. For instance, Low Rank Representation (LRR) \cite{8,9} based on nuclear norm represents the data points with the lowest-rank representation among all the candidates; Least Square Regression (LSR) \cite{10} with $\ell$$_2$ norm groups the highly correlated data; Exemplar-based Subspace Clustering (ESC) \cite{11} based on $\ell$$_1$ norm focuses on the class-imbalanced data; Elastic Net Subspace Clustering (ENSC) \cite{12} adopts both $\ell$$_1$ and $\ell$$_2$ norm to find better coefficients; Subspace Learning by $\ell$$_0$-Induced Sparisty \cite{13} employs proximal gradient descent to obtain a sub-optimal solution; while Sparse Subspace Clustering (SSC) \cite{14} with the $\ell$$_1$ norm calculates a sparse self-representation of data points. And Structured Sparse Clustering \cite{16} learns both affinity and segmentation of SSC.

SSC is inefficient when analyzing the large-scale dataset, so here come the Sparse Subspace Clustering by Orthogonal Matching Pursuit (SSC-OMP) \cite{17} and Subspace Clustering via Matching Pursuit (SSC-MP) \cite{18}, which are much faster than SSC.
However, SSC-OMP may suffer from a reduction of accuracy at the presence of noise\cite{19}.

In order to improve the accuracy of OMP, Active Orthogonal Matching Pursuit for Sparse Subspace Clustering (AOMP-SSC) \cite{20} updates the data and drops them randomly in the process of OMP, while Sparse Subspace Clustering by Rotated Orthogonal Matching Pursuit (SSC-ROMP) \cite{21} rotates the data points. Both AOMP-SSC and ROMP-SCC have raised the computational complexity of SSC-OMP, while the accuracy is not improved enough. Moreover, they both change the data points in the process of OMP, which disrupts the original distribution of the data points.

%
%
%

In this paper, we %
propose the concept of \emph{connection} and %
find that the number of connections of a data point is the potential sparsity of it when it comes to the affinity matrix, which affects the clustering result. Therefore, we propose a noise-robust Restricted Connection Orthogonal Matching Pursuit for Sparse Subspace Clustering (RCOMP-SSC) to address the issue of accuracy reduction of OMP and keep the original form of the data points, while still enjoying the low computational complexity. Also, we develop a framework of data point selection with control matrix of the inner products of the residuals and the other data points, on which we can realize RCOMP. More importantly, the control matrix is scalable, since most of the data point selection strategies, including AOMP-SSC and ROMP-SSC, can act on it. We demonstrate through experiments on different tasks of computer vision, i.e., clustering images of faces under varying illumination conditions (EYaleB) and hand written digit clustering (Usps) that RCOMP-SSC outperforms other methods in terms of accuracy and computational time.

\section{PRELIMINARY}
\label{sec:pre}

Given a dataset {\bf \emph{X}} = [{\bf\emph{x}}$_1$, {\bf\emph{x}}$_2$ ,..., {\bf\emph{x}}$_N$] $\in $ $\mathbb{R}$$^{D\times N}$, where every {\bf\emph{x}}$_\emph i$ is a data point of $\ell$$_2$ norm. D stands for the original dimension of the data. And there are totally N data points in the dataset. SC clusters data points into their original subspaces%
\{{\bf\emph S}$_\emph i$\}$^n_{\emph i = 1}$%
with the dimensions%
\{{\bf\emph d}$_\emph i$\}$^n_{\emph i = 1}$%
. That is,

\begin{equation} \label{eq1}
\setlength{\abovedisplayskip}{1pt}
\setlength{\belowdisplayskip}{1pt}
\bm x_{i}=\bm{Xc}_{i} \quad s.t.  c_{ii}=0
\end{equation}

\subsection{Sparse Subspace Clustering (SSC)}
\label{ssec:2.1}

SSC calculates the coefficient matrix %
{\bf\emph C}$_{SSC}$=\{{\bf\emph c}$_\emph i$\}$^N_{\emph i = 1}$ %
in (2) by $\ell$$_1$ norm, which aims at finding a sparse representation of each data point. Every {\bf\emph{c}}$_\emph i$ is correspond to {\bf\emph{x}}$_\emph i$ , of which the nonzero entries indicate the data points from the same subspace. And the last part measures the similarity between {\bf\emph{x}}$_\emph i$ and {\bf\emph{X}}{\bf\emph{c}}$_\emph i$  when there is noise.

\begin{equation} \label{eq2}
\setlength{\belowdisplayskip}{8pt}
\bm c_i=\mathop{argmin}_{\bm c_i}
    \|\bm c_i\|_1+
    \lambda \|{\bm x_i}-\bm{Xc}_{i}\|_2 \ s.t.c_{ii}=0
\end{equation}

After calculating the coefficient matrix, SSC applies the spectral clustering \cite{15} to the affinity matrix below:

\begin{equation} \label{eq3}
\bm W=|\bm C|+|\bm C|^T
\end{equation}

\subsection{Sparse Subspace Clustering by Orthogonal Matching Pursuit(SSC-OMP)}
\label{ssec:2.2}

SSC-OMP finds \emph k neighbors of {\bf\emph x}$_\emph i$ and represents {\bf\emph x}$_\emph i$ with its neighbors (4). In OMP, another data point \{{\bf\emph x}$_\emph j$ s.t. j$\neq$i\} that has the largest absolute inner product with the current residual will be added to the neighbor set of {\bf\emph x}$_\emph i$. Then {\bf\emph x}$_\emph i$ is iteratively projected onto the span of its current neighbors so as to update the residual ({\bf\emph {r}}) until \emph k neighbors of {\bf\emph x}$_\emph i$ is found or the residual is small enough.

\begin{equation} \label{eq4}
\bm c_i=\mathop{argmin}_{\bm c_i}
    \|{\bm x_i}-\bm{Xc}_{i}\|^2_2
    \ s.t. \ \|\bm c_i\|_0\leq k, \ c_{ii}=0
\end{equation}

\section{RCOMP-SCC}
\label{sec:RCOMP}
In this section, we first illustrate the inspiration behind our proposed algorithm. And then we demonstrate the detailed procedure of RCOMP-SSC. The analysis of the algorithm is showed at the last part.
\subsection{Intuition}
\label{ssec:3.1}

\begin{definition}[connection]
A connection is formed from {\bf\emph x}$_i$ to {\bf\emph x}$_j$ when {\bf\emph x}$_i$ choose {\bf\emph x}$_j$ as its neighbor.
\end{definition}

\noindent In OMP, the sparsity of the representation of {\bf\emph x}$_i$ is definite as the parameter \emph k is settled, which is exactly equal to \emph k. However, the clustering result relies on the structure of {\bf\emph W} in (3), which computes not only the connections between {\bf\emph x}$_i$ and its neighbors but also the connections between {\bf\emph x}$_i$ and \{{\bf\emph x}$_j$  s.t. j $\neq$ i\} who choose {\bf\emph x}$_i$ as a neighbor. We define the total number of the two kinds of connections of one data point as \emph{con}, and the sparsity of {\bf\emph x}$_i$ is represented by \emph{con} rather than \emph k when it comes to the affinity matrix. Notice that the total number of \emph{con} in affinity matrix is constant as equals to 2\emph{Nk}. However, the \emph{con} is not equal for each data point, since some data points have large ones for being chosen as neighbors more times. The other data points with small \emph{con} could not be clustered correctly. Therefore, it is necessary to have roughly equal \emph{con} among data points so as to get a correct clustering result of all data points.


Moreover, a subspace is grouped if any two data points in the subspace is connected by a couple of points which belong to the same subspace \cite{15}, which means a data point can be clustered correctly as long as it has a connection to the right group (one connection is enough). There are totally 2\emph{Nk} connections formed in the process of OMP, while not all of them are necessary or effective. Excessive connections of one point will weaken the \emph{space detection property} (SDP) \cite{22} of the similarity matrix, which leads to a decrease of the clustering accuracy. Therefore, we consider restricting the number of connections that each data point could have.

\subsection{Algorithm}
\label{ssec:3.2}

During the first iteration of OMP, we restrict the number of connections of each data point, and this algorithm is named as RCOMP. In practice, we only take \emph{connection}-1 (connection from {\bf\emph x}$_i$ to the first neighbor it chooses) and \emph{connection}-2 (connections from {\bf\emph x}$_i$ to \{{\bf\emph x}$_j$  s.t.j$\neq$i\} who choose {\bf\emph x}$_i$ as the first neighbor) as connections since they have the largest coefficients and the rest is insignificant in contrast. Discarding the connections with negligible coefficients relaxes the limit on selecting neighbors and reduces the computational time.
The whole procedure of RCOMP is shown in Algorithm 1.

\begin{algorithm}[t]
\setlength{\textfloatsep}{6pt}
\setlength{\floatsep}{20pt}
\caption{RCOMP-SCC}
\label{alg1}
{\bf Input:} Dataset {\bf \emph{X}} = [{\bf\emph{x}}$_1$, {\bf\emph{x}}$_2$ ,..., {\bf\emph{x}}$_N$] $\in $ $\mathbb{R}$$^{D\times N}$,
OMP iteration \emph k, the number of connections the data point could have \emph{rcon}.

\begin{algorithmic}[1]

\STATE Initialize {\bf\emph {ncon}} $\in\mathbb{R}^N$ with all elements equal to \emph{rcon}, residual {\bf\emph {r}} $\in\mathbb{R}^N$.
\FOR {i = 1, 2, . . . , N}
\STATE {\bf\emph {r}} = {\bf\emph{x}}$_i$.
\FOR {s = 1, 2, . . . , k}
\STATE Find the neighbors of {\bf\emph{x}}$_i$ by Algorithm 2.
\ENDFOR
\STATE Compute the {\bf\emph{c}}$_i$.
\ENDFOR
\STATE Set {\bf\emph {C}} = [{\bf\emph{c}}$_1$, {\bf\emph{c}}$_2$ ,..., {\bf\emph{c}}$_N$].
\STATE Compute {\bf\emph {W}} in (3).
\STATE Apply spectral clustering on {\bf\emph {W}}.

\end{algorithmic}
{\bf Output:} Clustering labels.
\end{algorithm}

\begin{algorithm}[t]

\caption{Control Matrix Framework of RCOMP}
\label{alg2}
{\bf Input:} {\bf\emph{x}}$_i$ $\in $ $\mathbb{R}$$^{D}$, {\bf \emph{X}} $\in $ $\mathbb{R}$$^{D\times N}$, {\bf\emph {ncon}} $\in\mathbb{R}^N$, the current iteration \emph k, residual {\bf \emph{r}} $\in\mathbb{R}^D$.

\begin{algorithmic}[1]
\STATE Initialize {\bf\emph{dot}} = {\bf\emph{x}}$_i$ * {\bf \emph{X}}$^T$, the control matrix {\bf\emph M} $\in\mathbb{R}^{N\times N}$ with all elements equal to 1, while the diagonal is 0.
\STATE Update {\bf\emph{dot}} with (5).
\STATE Find {\bf\emph{x}}$_j$ as the neighbor of {\bf\emph{x}}$_i$, {\bf\emph{x}}$_j$ = argmax \emph{dot}$_j$.
\IF {\emph k == 1}
\STATE \emph{ncon}$_j$ $\leftarrow$ \emph{ncon}$_j$ - 1.
\STATE \emph M$_{ji}$ = 0.
\IF {\emph{ncon}$_j$ == 0}
\STATE Set the \emph jth column of {\bf\emph M} to be 0.
\ENDIF
\ENDIF
\STATE Update {\bf\emph{r}} with OMP.
\end{algorithmic}

{\bf Output:} {\bf\emph{x}}$_j$, the updated {\bf\emph M}, {\bf\emph{ncon}} and {\bf\emph {r}}.
\end{algorithm}
\begin{definition}[control matrix]
Given a dataset {\bf \emph{X}} = {\rm [}{\bf\emph{x}}$_1$, {\bf\emph{x}}$_2$ ,..., {\bf\emph{x}}$_N${\rm ]} $\in $ $\mathbb{R}$$^{D\times N}$, control matrix {\bf\emph M} $\in $ $\mathbb{R}$$^{N\times N}$, M$_{ij}$ is the control coefficient which is applied to multiplying the inner products of {\bf\emph x}$_i$ (or its residuals) and {\bf\emph x}$_j$(s.t.j$\neq$i) so as to control the neighbor selection.
\end{definition}

OMP selects {\bf\emph x}$_j$ that has the largest absolute inner product with {\bf\emph x}$_i$ as the first neighbor of {\bf\emph x}$_i$, so that we can control the selection by controlling the inner product of {\bf\emph x}$_i$ and \{{\bf\emph x}$_j$ s.t. j$\neq$i\} (Algorithm 2). We present a N$\times$N control matrix ({\bf\emph M}) to realize RCOMP by multiplying the inner products ({\bf\emph{dot}}) with the corresponding entries of {\bf\emph M} as shown in (5). The number of connections of {\bf\emph x}$_j$ could still have (\emph{ncon$_j$}) decreases when a new connection formed (step 5). When \emph{ncon$_j$} reduces to 0, which means the number of connections of {\bf\emph x}$_j$ reaches the limit, the {\bf\emph x}$_j$ could not be chosen as the neighbor of the other data points anymore(step 8).

\begin{equation} \label{eq5}
\bm {dot}'=\{
    \bm {dot}_i' \
    | \ dot'_{ij}=dot_{ij} \times M_{ij}
\}
\end{equation}

\begin{figure}[t]
\setlength{\abovecaptionskip}{0pt}
\setlength{\belowcaptionskip}{0pt}

\begin{minipage}[b]{.48\linewidth}
  \centering
  \centerline{\includegraphics[width=2.2cm]{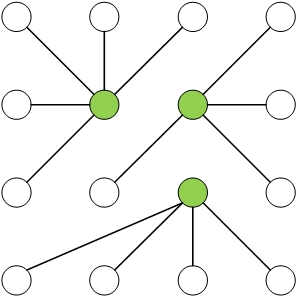}}
  \vspace{0.3cm}
  \centerline{(a) OMP}
\end{minipage}
\hfill
\begin{minipage}[b]{0.48\linewidth}
  \centering
  \centerline{\includegraphics[width=2.2cm]{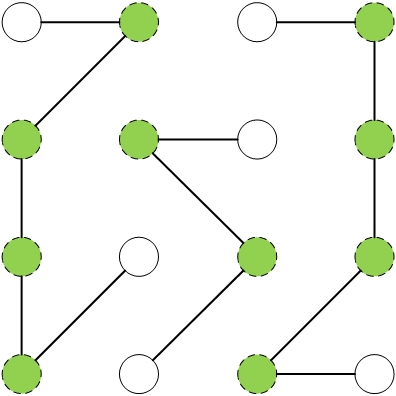}}
  \vspace{0.3cm}
  \centerline{(b) RCOMP}
\end{minipage}

\caption{Example of connections between data points}
\label{fig:conn}
\end{figure}

\subsection{Analysis}
\label{ssec:3.3}

The sparse solution is not unique since any \emph d$_i$ linearly independent points from \emph S$_i$ can represent a point {\bf\emph x}$_i$$\in${\emph S}$_i$ \cite{17}. Therefore, we can still get a sparse representation of {\bf\emph x}$_i$ even if we change the points that OMP chooses.

If there are enough candidates for the data points behind to choose as neighbors, the restriction of RCOMP would not worsen the clustering result. As a consequence, the number of connections each data point could have (\emph{rcon}) is related to the size of the dataset. The larger the dataset, the smaller the \emph{rcon}. In particular, when \emph{rcon} is equal to 2, the RCOMP would form a linear structure among data points (Fig.2.(b)). Generally, we set \emph{rcon} = 2, while \emph{rcon} = 3 for small datasets in extreme situation.

We can get a roughly equal \emph{con} among data points by restricting it to \emph{rcon} since the total number of \emph{con} is instant, which increases the number of connections of data points whose \emph{con} is relatively small in OMP. Therefore, all the data points in {\bf\emph X} can be clustered correctly. As shown in Fig.2., in the case where the total number of connections of all data points is the same, RCOMP tends to own more data points whose connections is more than one (green points). And the dotted points mean they can¡¯t be connected again.


\begin{equation} \label{eq6}
\setlength{\belowdisplayskip}{12pt}
\begin{aligned}
\{
    \bm {x}_j \ |
    \exists \bm r_i \in S_i,\bm x_j \notin S_i, \forall \bm x_i \in S_i ,|\bm r_i\bm x_i|<|\bm r_i\bm x_j|
\}
\end{aligned}
\end{equation}

Furthermore, when the subspaces are not disjoint (or there is noise), OMP may forms too many wrong connections with data points ({\bf\emph X}$_0$ =\{{\bf\emph x}$_j$\} in (6)) which are near the intersections between subspaces. In contrast, RCOMP allows only 2 (in general) connections of one data point, so that there would be at most 2 wrong connections of each data point in {\bf\emph X}$_0$. Therefore, the restriction of RCOMP makes it robust to the noise.

Notice that OMP selects neighbors by the absolute value of inner products, which is updated by {\bf\emph{M}}. Therefore, most of the data point selection strategies can be implemented on the framework of control matrix {\bf\emph{M}} simply by updating the control coefficients of it. For instance, setting the \emph ith column of {\bf\emph{M}} to be zero actually drops the \emph ith data point just as AOMP-SSC dose.

\section{EXPERIMENTS}
\label{sec:experiment}


\subsection{Synthetic Experiment}
\label{ssec:4.1}

\begin{figure}[t]
\setlength{\abovecaptionskip}{0.cm}
\setlength{\belowcaptionskip}{-0.cm}

\begin{minipage}[b]{.48\linewidth}
  \centering
  \centerline{\includegraphics[width=4cm]{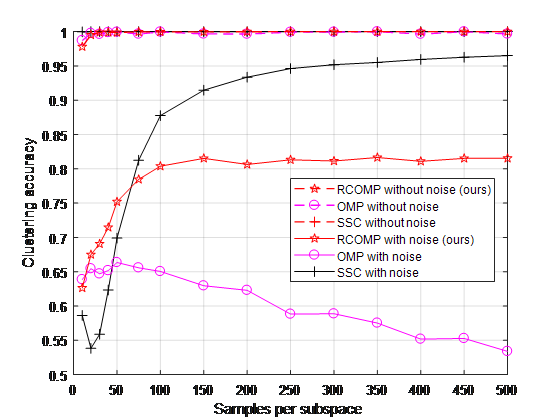}}
  \vspace{0.25cm}
  \centerline{(a) Accuracy}
\end{minipage}
\hfill
\begin{minipage}[b]{0.48\linewidth}
  \centering
  \centerline{\includegraphics[width=4cm]{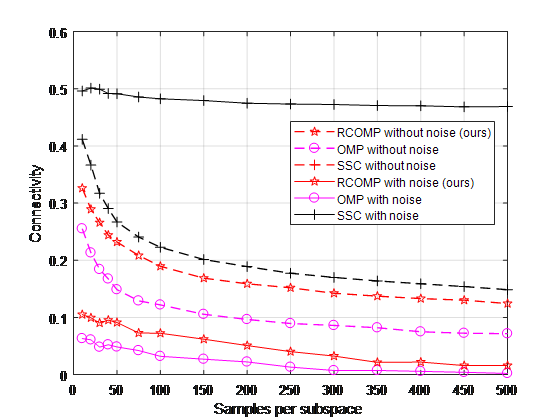}}
  \vspace{0.25cm}
  \centerline{(b) Connectivity}
\end{minipage}
\begin{minipage}[b]{.48\linewidth}
  \centering
  \centerline{\includegraphics[width=4cm]{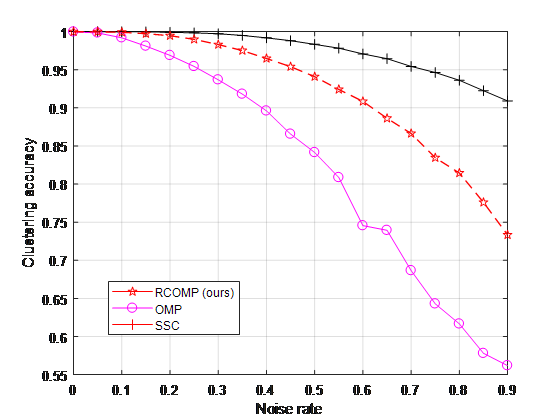}}
  \vspace{0.25cm}
  \centerline{(c) Noise Rate Accuracy}
\end{minipage}
\hfill
\begin{minipage}[b]{0.48\linewidth}
  \centering
  \centerline{\includegraphics[width=4cm]{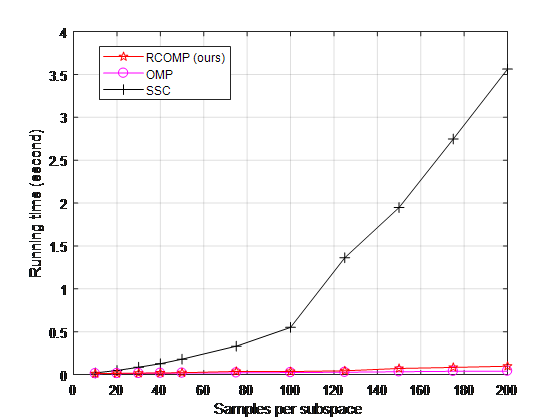}}
  \vspace{0.25cm}
  \centerline{(d) Computation Time}
\end{minipage}

\caption{Experiments on synthetic data}
\label{fig:syn_data}
\end{figure}

We randomly generate \emph n $=$ 3 subspaces of dimension \emph d $=$ 6 in 40-D ambient space. There are 10 to 500 data points in every subspace. The noise rate is 0.8 in Fig.2.(a). We set up \emph{rcon} $= $ 2.
The average results of 100 trials are shown in Fig.2.

All of the three algorithms display good performance when there is no noise (only under this experiment settings). And RCOMP exceeds OMP at the presence of noise in Fig.2(a).
More importantly, as data density grows, there are more and more data points near the intersections of subspaces. As a consequence, OMP suffers from an accuracy decline, while RCOMP keeps good performance. Therefore, we come to a conclusion that RCOMP is more robust than OMP in large and noisy datasets, which will be verified again in the following two experiments. Similar results in connectivity \cite{23} is shown in Fig.2(b).

In order to test the performance of RCOMP under different rates of noise, we set noise rates various from 0 to 0.9 with 200 data points of each subspace. Other settings are the same as the above experiment. Fig.2 (c) shows that RCOMP outperforms OMP in various rates of noise.

Although SSC outperforms RCOMP in terms of accuracy and connectivity, RCOMP is several magnitudes faster than SSC (Fig.2(d)).  This is in line with the intention that RCOMP aims at keeping a balance between accuracy and efficiency. Moreover, RCOMP achieves better performance over SSC in the next two real-world datasets.

\subsection{Handwritten Digits Images Segmentation}
\label{ssec:4.2}

We use Usps, a handwritten 0-9 digits database of 8-bit grayscale images, to test RCOMP in contrast to other methods. There are 1100 images for each digit, from which we randomly choose $\{$200, 400, 600, 800, 1000$\}$. All of the image are projected to dimension 200 using PCA. We set iteration \emph k $=$ 3, \emph{rcon} $=$ 2 since the dataset is large enough, and the number of clusters we choose is 5. After 100 times trials, the result of accuracy and computational time is shown in Table.1. RCOMP outperforms the others in terms of accuracy in all five groups, which shows the robustness of RCOMP in various sizes of datasets. Moreover, RCOMP is significantly faster than the other clustering methods (except for OMP), which makes it preferable for large-size tasks.

\subsection{Face Image Clustering}
\label{ssec:4.3}

In this experiment, we test our algorithm on the Extended Yale B dataset \cite{24}, which consists of frontal face images of 38 persons under 64 different illumination conditions. The size of each image is down sampled to 48$\times$42 pixel. We randomly pick \emph n $\in$ $\{5$, 10, 20, 30, 38$\}$ persons with all the images of each. We set the iteration time \emph k $=$ 5, \emph{rcon} $=$ 3 since there are only 64 data points of each subspace. The result is shown in Table.2 with 100 trials of each \emph n, which demonstrates the superiority of RCOMP in multiple classes.

\begin{table}[t]
\setlength{\abovecaptionskip}{2pt}
\setlength{\belowcaptionskip}{5pt}
\centering
\caption{Accuracy (\%)/Computational time (s) on Usps}
\vspace{0.3cm}

\begin{tabular}{c|ccccc}
\hline
&200&400&600&800&1000 \\ \cline{1-6}

RCOMP&\bf{69.56}/&\bf{72.37}/&\bf{72.08}/&\bf{71.77}/&\bf{70.00}/ \\
(ours)&0.22&0.60&1.27&2.66&5.02 \\

OMP&62.52/&61.21/&56.17/&55.65/&55.79/ \\
&0.08&0.18&0.33&0.53&0.75 \\

AOMP&67.75/&70.20/&66.62/&67.84/&67.65/ \\
&1.34&5.04&11.4&20.5&32.0 \\

ROMP&68.10/&69.90/&66.92/&66.24/&66.37/ \\
&2.02&6.92&15.8&27.9&43.9 \\

SSC&61.24/&60.28/&60.81/&60.72/&61.36/ \\
&6.54&37.8&113&258&515 \\

ENSC&59.56/&57.97/&54.97/&52.75/&52.22/ \\
&37.4&98.4&106&243&328 \\

LRR&62.04/&64.54/&63.84/&65.30/&66.68/ \\
&5.86&14.7&31.4&56.6&101 \\

\hline
\end{tabular}
\end{table}
\begin{table}[t]
\centering
\caption{Accuracy (\%) on EYaleB}
\vspace{0.3cm}

\begin{tabular}{c|ccccc}
\hline
&5&10&20&30&38 \\ \cline{1-6}

RCOMP(ours)&\bf{96.67}&\bf{94.77}&\bf{89.20}&\bf{86.00}&\bf{83.33} \\

OMP&95.24&86.77&81.57&77.60&76.59 \\

AOMP&95.23&88.15&82.26&79.58&77.96 \\

ROMP&96.52&93.94&85.92&81.29&77.96 \\

SSC&83.13&58.31&56.36&54.75&57.66 \\

ENSC&76.47&76.19&64.74&60.56&56.53 \\

LRR&67.37&76.60&69.50&66.26&64.08 \\

\hline
\end{tabular}
\end{table}
\section{CONCLUSIONS}
\label{sec:conclusion}



We propose a noise-robust RCOMP-SSC algorithm, which obtains a roughly equal number of connections of each data point and alleviates the effects of data points near the intersections of subspaces by restricts the number of connections of each data point. And it is realized on the scalable framework of control matrix we develop. We illustrate the inspiration behind our algorithm and present theoretic analysis of it. %
Experiments on synthetic data and different tasks of computer vision show that our algorithm outperforms other clustering methods in terms of accuracy, computational time and scalability in various sizes and classes of datasets.

%

\bibliographystyle{IEEEbib}
\bibliography{strings}

\end{document}